\documentclass[12pt,letterpaper]{article}
\usepackage[a4paper, total={7in, 10in}]{geometry}

\usepackage{graphicx}
\usepackage{helvet}
\usepackage{authblk}
\usepackage{amsmath} 
\usepackage{amssymb} 
\usepackage{orcidlink} 
\usepackage[super,comma,sort&compress]  
   {natbib}
\usepackage[right]{lineno} \nolinenumbers
\usepackage{color}
\definecolor{darkred}{rgb}{0.9,0,0.3}
\definecolor{darkblue}{rgb}{0,0.3,0.9}

\definecolor{vdarkred}{rgb}{0.7,0,0.2}
\definecolor{vdarkblue}{rgb}{0,0.2,0.7}

\definecolor{Slc17a6}{rgb}{0, 1, 0.82}
\definecolor{Slc17a7}{rgb}{0.69, 1, 0}
\definecolor{Nr4a2}{rgb}{1, 0, 0.4}
\definecolor{Th}{rgb}{1, 0.4, 0}

\makeatletter
\renewcommand{\maketitle}{\bgroup\setlength{\parindent}{0pt}
\begin{flushleft}
  \textbf{\@title}
  
  \@author
\end{flushleft}\egroup}
\makeatother

\title{Tera-MIND: Tera-scale mouse brain simulation via spatial mRNA-guided diffusion}
\date{}

\author[1,\orcidlink{0000-0002-6898-8698}]{Jiqing Wu}
\author[2, \orcidlink{0000-0002-7058-1969}]{Ingrid Berg}
\author[3, 4, \orcidlink{0000-0002-8948-7892}]{Yawei Li}
\author[3, \orcidlink{0000-0002-2542-3611}]{Ender Konukoglu}
\author[1,5,\orcidlink{0000-0001-9206-4885}, *]{Viktor H. Koelzer}

\affil[1]{Department of Biomedical Engineering, University of Basel, Switzerland}
\affil[2]{Department of Pathology and Molecular Pathology, University Hospital, University of Zurich, Switzerland}
\affil[3]{Computer Vision Lab, ETH Zurich, Switzerland}

\affil[4]{Integrated System Laboratory, ETH Zurich, Switzerland}
\affil[5]{Institute of Medical Genetics and Pathology, University Hospital Basel, Switzerland}

\affil[*]{Correspondence: Viktor.Koelzer@usb.ch}

\begin{document}

\maketitle

\section*{SUMMARY}
Holistic 3D modeling of molecularly defined brain structures is crucial for understanding complex brain functions. Using emerging tissue profiling technologies, researchers charted comprehensive atlases of mammalian brain with sub-cellular resolution and spatially resolved transcriptomic data. However, these tera-scale volumetric atlases pose computational challenges for modeling intricate brain structures within the native spatial context. We propose \textbf{Tera-MIND}, a novel generative framework capable of simulating \textbf{Tera}-scale \textbf{M}ouse bra\textbf{IN}s in 3D using a patch-based and boundary-aware \textbf{D}iffusion model. Taking spatial gene expression as conditional input, we generate virtual mouse brains with comprehensive cellular morphological detail at teravoxel scale. Through the lens of 3D \textit{gene}-\textit{gene} self-attention, we identify spatial molecular interactions for key transcriptomic pathways, including glutamatergic and dopaminergic neuronal systems. Lastly, we showcase the translational applicability of Tera-MIND on previously unseen human brain samples. Tera-MIND offers an efficient generative modeling of whole virtual organisms, paving the way for integrative applications in biomedical research.
\section*{KEYWORDS}


Spatial transcriptomics, mouse and human brain, Generative AI, diffusion model

\section*{INTRODUCTION}

Recently, a network of researchers~\cite{yao2023high,zhang2023molecularly,langlieb2023molecular,shi2023spatial} has reached a groundbreaking milestone by charting the first complete cellular atlas of the adult mouse brain, the most extensively studied animal model in neuroscience. Using advanced spatial transcriptomics methods~\cite{bressan2023dawn,moses2022museum,chen2015spatially} capable of profiling molecularly defined brain anatomy and morphology at sub-cellular resolution, the collective studies established foundational resources and datasets, providing an unprecedented framework for systematically exploring the molecular-driven spatial organization of the mammalian brain. Consequently, this landmark achievement opens the door for next-generation investigations into intricate brain functions characterized by exceptional spatial complexity. However, these tera-scale volumetric datasets, including spatial molecular data (\textit{e.g.,} mRNA readouts acting as proxies for gene expression) and paired morphological bioimages, pose substantial computational challenges~\cite{wu2024towards} for \textit{in-silico} modeling of the mouse brain. Integrating spatial gene expression with 3D brain morphology to uncover functional relationships at whole organ scale remains an open challenge.

Meanwhile, generative artificial intelligence (GenAI) has emerged as a transformative tool for automating visual content creation and manipulation~\cite{cao2024survey,zeng2024dawn}. Given a text prompt that describes the desired visual content, GenAI models~\cite{goodfellow2014generative,ho2020denoising} are able to produce highly realistic images and videos~\cite{rombach2022high,peebles2023scalable} with remarkable efficiency. 
In biomedical research, the applications of GenAI have gained domain-specific recognition in various areas such as digital histopathology~\cite{alajaji2023generative}, drug screening~\cite{wu2024gilea,lamiable2023revealing}, and molecule design~\cite{cao2024survey}. However, the direct application of existing GenAI methods to whole organ simulation has currently remained infeasible due to technological limitations. Developed upon the framework of Generative Adversarial Nets (GAN)~\cite{goodfellow2014generative}, the GigaGAN~\cite{kang2023scaling} model was trained on 96-128 A100 GPUs to synthesize $4,096 \times 4,096$ color images. Using up to 6,144 H100 GPUs, Movie Gen~\cite{polyak2024movie}, a diffusion-based model~\cite{ho2020denoising}, achieved video generation results with $1,092 \times 1,080 \times 256$ spatial-temporal resolution. By contrast, each whole slide image (WSI; $0.108 \mu m/\mathrm{pixel}$) of the quality-controlled mouse brain atlas has a spatial resolution of $73,216 \times 105,984$. Consequently, stacking 50 slices of these DAPI (4',6-diamidino-2-phenylindole) and PolyT-stained (total mRNA signals) WSIs results in $0.77\times 10^{12}$ voxels in total. This immense data scale of WSIs and paired gene expression arrays significantly intensifies the computational demands associated with GenAI modeling of a virtual mouse brain.

Addressing memory bottlenecks in high-resolution image generation has previously motivated the development of patch-based approaches~\cite{lin2019coco,lin2021infinitygan,xu2021positional,nikankin2023sinfusion,ding2023patched}, which enforce boundary consistency between generated neighboring patches. 
Similar to many standard GenAI models, these approaches can also leverage the interplay between textual and visual modalities and create desired visual content conditioned on text prompts. While this design choice has shown remarkable success in natural image synthesis and artistic content creation, it is suboptimal for biomedical applications -- particularly for the accurate 3D molecular-driven reconstruction of mammalian brains. The inherent complexity of brain morphology controlled by spatial molecular interactions demands a generative framework that can capture molecular-to-morphology spatial associations with precision. In response to these challenges, we propose \textbf{Tera-MIND}, a novel GenAI approach designed for the paired data structure of spatial mRNA arrays (gene expression, prompts) and WSIs (brain morphology, bioimages).
Specifically, Tera-MIND employs a newly introduced patch-based and boundary-aware diffusion model, which allows the reconstruction of tera-scale mouse brains with high fidelity. Owing to the patch-wise training and inference paradigm, Tera-MIND models \textit{in-silico} mammalian brain(s) with computational efficiency. As a result, the whole simulation process can be efficiently executed on a single DGX A100 machine, significantly lowering hardware demands while maintaining scalability for tera-scale data processing. This efficiency underscores the potential wide applicability of Tera-MIND for biomedical applications. Our contributions can be summarized as follows:

\begin{itemize}
\item By conditioning on 3D spatial gene expression as the input prompt, Tera-MIND enables the seamless generation of \textit{in-silico} mouse brains at the scale of $0.77$ teravoxels. Moreover, our approach accurately preserves fine-grained morphological structures across cellular, tissue, and slice-wise scales, supporting detailed exploration and comparison of brain architecture.
\item Leveraging 3D \textit{gene}-\textit{gene} self-attention mechanisms, we quantify and visualize spatial molecular interactions of key pathways that contribute to fundamental brain functions, including those involved in glutamatergic (\textit{\textcolor{Slc17a6}{Slc17a6}} and \textit{\textcolor{Slc17a7}{Slc17a7}}) and dopaminergic (\textit{\textcolor{Nr4a2}{Nr4a2}} and \textit{\textcolor{Th}{Th}}) neuronal systems.
\item Similar to the translational aims of wet-lab mouse brain studies, we show that Tera-MIND's simulation results are applicable to previously unseen human health and glioblastoma brain samples, underscoring its potential in advancing the understanding of complex brain functions and diseases for real-world biomedical applications.
\end{itemize}

\section*{RESULTS}

\subsection*{Model overview of Tera-MIND}
To resolve the high-volume computational demand, we develop a novel patch-based and boundary-aware diffusion model \textbf{Tera-MIND}, designed for the scalable and seamless generation of teravoxel mouse brains (See Fig.~\ref{Fig1} (a, b)). Considering the spatial linkage of gene expression variability and brain morphology, we propose to modulate a 3D \textit{gene}-morph (gm) UNet using a tailored 3D \textit{\textit{gene}-\textit{gene}} (gg) block. By feeding the spatial gene expression array (input prompt) to the 3D-gg block, we achieve the sub-cellular control of brain morphology generation through the 3D-gm UNet, please see Fig.~\ref{Fig1} (c) for more illustrations. To assess the role of individual genes and their relationships within gene groups, the proposed 3D \textit{\textit{gene}-\textit{gene}} block includes a 3D self-attention layer that learns the \textit{\textit{gene}-\textit{gene}} interaction level within the native 3D spatial context. Following the attention architecture introduced in Diffusion Transformer (DiT)~\cite{peebles2023scalable}, we then inject learned \textit{\textit{gene}-\textit{gene}} attention representations into the 3D \textit{gene}-morph cross-attention block, eventually modulating brain morphology generation using 3D-gm UNet. As a result, the output of Tera-MIND is a stack of neigboring DAPI- and PolyT-stained image tiles that capture brain morphology. In addition to the patch-wise reconstruction path (black arrows in Fig.~\ref{Fig1} (c, d)), we further introduce a boundary-aware path (gray arrows in Fig.~\ref{Fig1} (c, d), please see also~\cite{ding2023patched}) to output center-cropped brain morphology patches. Specifically, the training process involves randomly cropping paired $n$-plex gene expression arrays and their corresponding morphological bioimages from training mouse brain atlas(es)~\cite{yao2023high}.
Our patch-based approach thus enables efficient training of the Tera-MIND model on 2 $\times$ 40GB A100 GPUs. During inference, Tera-MIND supports the seamless generation of unseen mouse brain(s) with high fidelity, while the hardware requirement remains moderate, \textit{e.g.}, 7 days on a single DGX machine with 8$\times$ 40GB A100 GPUs. For model details we refer interested readers to the \hyperref[sec:Methods]{STAR METHODS} section and code repository.

\subsection*{Tera-MIND accurately generated tera-scale mouse brain(s) by spatial gene expression.}

Consistent with the primary analysis described in~\cite{zhang2023molecularly}, which was conducted on a comprehensive brain atlas of a P56 female adult mouse (Sampleid: 638850), we present \textbf{main} simulation results of Tera-MIND based on the same brain atlas.  The holistic 3D comparison between the ground truth (GT) and our generation result is illustrated in Fig.~\ref{Fig1} (b) and Fig.~\ref{Fig2} (a). Given 3D spatial gene expression data as an input, Tera-MIND achieves the faithful and comprehensive reconstruction of 3D brain morphology, as captured by a sequential stack of \textcolor{blue}{DAPI} and \textcolor{green}{PolyT}-stained WSIs.  Specifically, the generated mouse brain has the same spatial resolution of 0.77 teravoxels as GT. At the level of individual WSIs (Fig.~\ref{Fig2} (b, c)), Tera-MIND exhibits high-quality generation across multiple scales, including cellular, regional, and slice-wise morphological organization. Within distinct brain regions, such as the olfactory areas, isocortex, and cerebellar cortex, Tera-MIND accurately reconstructs complex morphological structures while preserving native cellular distributions. These results highlight the robustness of Tera-MIND for reconstructing 3D brain morphology directly from spatial gene expression profiles.

Complementary to the side-by-side qualitative comparisons, we perform thorough quantitative analyses of the reconstruction quality by Tera-MIND. For a fair and systematic assessment of the biomedical-specific outputs, we report not only commonly used metrics such as Peak Signal-to-Noise Ratio (PSNR) and Structural Similarity Index Measure (SSIM), but also include patch-based morphometric analyses of nuclear size and cell number. These domain-specific metrics are critical for evaluating the biological fidelity of reconstructed brain morphology at the cellular level. Then, we perform quanlitative (Fig.~\ref{Fig2} (d)) and quantitative (Fig.~\ref{Fig2} (e)) evaluation on Tera-MIND in comparison to state-of-the-art (SOTA) methods CoCoGAN~\cite{lin2019coco}, InfinityGAN~\cite{lin2021infinitygan}, IST-editing~\cite{wuediting}, SinFusion~\cite{nikankin2023sinfusion}, Patch-DM~\cite{ding2023patched}, \textit{etc}. Overall, Tera-MIND achieved superior performance in terms of better PSNR and SSIM scores. Furthermore, experimental results reported in Fig.~\ref{Fig2} (e) demonstrated marginal morphometric discrepancies between GT cellular patches and those generated by Tera-MIND, which are consistent with the results of stratified patch examples shown in Fig.~\ref{Fig2} (e).  Using spatial gene expression as the input prompt, these analyses confirm that Tera-MIND reliably reconstructs brain morphology across multiple scales, from cellular and regional structures to whole-brain organization. This consistent performance across scales highlights the efficacy of our approach for accurately capturing the complex spatial patterns of 3D brain morphology.

\subsection*{Tera-MIND identified biologically relevant \textit{gene}-\textit{gene} interactions in native 3D space.}

To explore the role of individual genes and their interactions in governing the reconstruction of the 3D mouse brain, we analyze the learned 3D \textit{gene}-\textit{gene} attention map and focus on two critical neuronal systems: glutamatergic (GLUT) and dopaminergic (DOPA) signaling pathways. These examples demonstrate that Tera-MIND is capable of capturing biologically relevant spatial molecular interactions that are essential for the structural and functional interplay of mammalian brains. 

\subsubsection*{GLUT neuronal system}
Glutamate is the primary excitatory neurotransmitter in the brain and plays a pivotal role in a wide range of brain functions~\cite{vandenberg2013mechanisms}. The vesicular glutamate transporters VGLUT1 and VGLUT2, encoded by \textit{Slc17a7} and \textit{Slc17a6} resp., regulate extracellular glutamate concentrations and synaptic signaling~\cite{vandenberg2013mechanisms,wallen2010genetic}. Moreover, the expression level of VGLUTs impacts the amount of glutamate loaded into vesicles and released, thereby affecting neurotransmission~\cite{wojcik2004essential}. Changes in VGLUT expression levels are associated with various neurological pathologies such as schizophrenia, neuropathic pain, and ischemia~\cite{moechars2006vesicular,sanchez2010transient}.
In adult mouse brains, spatial expression patterns of VGLUT1 and VGLUT2 are generally complementary but not exclusive ~\cite{wojcik2004essential}. 
Specifically, \textit{Slc17a7} (VGLUT1) is broadly expressed in the cerebral cortex, cerebellar cortex, hippocampus, and thalamus, \textit{etc}., while \textit{Slc17a6} (VGLUT2) can be found in subcortical regions such as the thalamus and spinal cord.

As shown in Fig.~\ref{Fig3} (a, left) and (b), spatial expression patterns of these transporters are accurately mapped in 3D space and align with prior studies~\cite{yao2023high, zhang2023molecularly}, which justifies the registration process of brain atlas utilized in this study. Interestingly, the 3D \textit{gene}-\textit{gene} attention map (Fig.~\ref{Fig3} (a, middle)) learned by Tera-MIND reveals widespread and heterogeneous attention signals (Fig.~\ref{Fig3} (a, right)) for the spatial interaction between \textit{Slc17a6} and \textit{Slc17a7}. Such heterogeneous attention levels can also be observed by the marginal distribution reported in Fig.~\ref{Fig3} (c), where the linear regression analysis identifies a strong correlation between expression and attention levels.
Since both transporters collectively contribute to the regulation of synaptic vesicle glutamate content in those overlapped regions, these heterogeneous spatial attention signals (Fig.~\ref{Fig3} (c, d)) likely encapsulate their accumulative impact on regional brain morphology and neurotransmission.

\subsubsection*{DOPA neuronal system}
In dopamine-mediated neuronal signaling, \textit{Nr4a2} (Nurr1) is an important transcription factor that is expressed in several regions of the central nervous system including the olfactory bulb~\cite{jakaria2019molecular}, the substantia nigra (SN) and ventral tegmental area (VTA). \textit{Nr4a2} plays a pivotal role in dopaminergic neurons by regulating several key genes (\textit{e.g.,} \textit{Th}) involved in dopamine synthesis, storage, and release. \textit{Th} encodes tyrosine hydroxylase, the rate-limiting enzyme in dopamine synthesis~\cite{jakaria2019molecular}. Both \textit{Nr4a2} and \textit{Th} genes are present in areas such as SN, VTA, and olfactory bulb of the mouse brain~\cite{roostalu2019quantitative,fu2012cytoarchitectonic}. 

The co-localization patterns have been correctly captured in spatial visualizations presented in Fig.~\ref{Fig3} (e, left) and (f), which are subsequently used to derive spatial attention levels using self-attention mechanism (Fig.~\ref{Fig3} (e, middle)). Reduced \textit{Nr4a2} and \textit{Th} expression are associated with Parkinson's disease~\cite{kalia2015parkinson,deng2020lipopolysaccharide}, a common neurodegenerative disorder with loss of dopaminergic neurons in the SN, leading to dopamine deficiency and motor function impairments~\cite{kalia2015parkinson,deng2020lipopolysaccharide}. This has also been observed in mice after deletion of \textit{Nr4a2} in mature dopaminergic neurons~\cite{decressac2013nurr1}. Though the linear regression analysis of Fig.~\ref{Fig3} (g) illustrates overall lower attention levels than the ones reported for GLUT system, we observe highly aggregated \textit{Nr4a2}-\textit{Th} attention signals in the olfactory bulb, SN and VTA (Bottom of midbrain label in Fig.~\ref{Fig3} (h)),  where both genes play a central role in regulating dopaminergic functions. As shown in Fig.~\ref{Fig3} (e, right) and (h), the 3D \textit{gene}-\textit{gene} attention map uncovers a spatially aligned interaction between \textit{Nr4a2} and \textit{Th} that reinforces their roles in the molecular regulation of dopaminergic pathways and cellular anatomical structures.

\subsection*{Tera-MIND achieved reproducible and robust results on three tera-scale mouse brains.}
In addition to the main mouse brain discussed above, two additional P56 adult mouse brains -- one male and one female -- were profiled in previous studies~\cite{yao2023high,zhang2023molecularly} for supporting analyses, as illustrated in Fig.~\ref{Fig4} (a) (middle and right). To rigorously evaluate the reproducibility and robustness of our findings, we extend the same simulation experiments conducted on the \textbf{main} mouse brain to both supporting mouse brains (\textbf{supp (m)} and \textbf{supp (f)}).

\noindent\textbf{Morphology generation}: As illustrated in Fig.~\ref{Fig4} (a, b), the side-by-side visual comparisons between our generated and GT results demonstrate convincing reconstructions of brain morphology on all three tera-scale mouse brains. Consistent with the main generation results (Fig.~\ref{Fig1} (b) and the left plots of Fig.~\ref{Fig4} (a, b)), our supporting experiments highlight competitive and reliable 3D and slice-wise image generation quality, as elaborated by the holistic and multi-scale visualization shown in the middle and right plots of Fig.~\ref{Fig4} (a, b). For quantitative evaluation, Tera-MIND is benchmarked against seven SOTA methods COCO-GAN~\cite{lin2019coco}, InfinityGAN~\cite{lin2021infinitygan}, MS-PIE~\cite{xu2021positional}, SST-editing~\cite{wu2024sst}, IST-editing~\cite{wuediting}, SinFusion~\cite{nikankin2023sinfusion}, and Patch-DM~\cite{ding2023patched}. Complementary to PSNR and SSIM scores, we incorporate the spatial Fr\'echet Inception Distance (sFID)~\cite{nash2021generating}, which prioritizes spatially relevant evaluations by rewarding image distributions that preserve coherent spatial structures. Across all three brain instances, Tera-MIND consistently outperforms competing methods, in terms of achieving better PSNR, SSIM, and sFID scores (Fig.~\ref{Fig4} (c)). When evaluated on domain-specific metrics such as cell number and nuclear size (Fig.~\ref{Fig4} (d, e)), Tera-MIND yields minimal distributional discrepancies relative to GT. This highlights its capacity to maintain high-fidelity not only in structural reconstruction but in the reproduction of biologically relevant features.

\noindent\textbf{Pathway identification}: In the context of the GLUT pathway, Fig.~\ref{Fig5} (a, b) present a highly consistent and spatially comparable localization of \textit{Slc17a6} and \textit{Slc17a7} expression patterns across two independent supporting instances. These expression maps are accurately registered in 3D space and align well with the ones from our main results and previous studies~\cite{yao2023high,zhang2023molecularly}. Similarly as observed in the main results, Fig.~\ref{Fig5} (c) and (d) illustrate that Tera-MIND is capable of learning strong and heterogeneous \textit{Slc17a6}-\textit{Slc17a7} attention levels driven by the positively correlated gene expression levels. This suggests their synergistic and collective roles in regulating neurotransmission and cellular morphology across overlapped regions. 
Regarding the DOPA pathway, despite the absence of the olfactory bulb in the raw data of the supporting mouse brains, the gene expression patterns of \textit{Nr4a2} and \textit{Th} exhibit a comparable emergence in the SN and VTA regions (See also Fig.~\ref{Fig5} (e, f)). Consistent with main results, the linear regression analysis of Fig.~\ref{Fig5} demonstrates the comparable correlation between gene expression and attention levels. Moreover, we witness reproducible and clear identification of \textit{Nr4a2}-\textit{Th} attention levels derived from Tera-MIND in the same regions for both supporting mouse brains (Fig.~\ref{Fig5} (e, h)). These results confirm a consistent expression pattern of critical genes associated with dopaminergic function in these specific regions. Please see also the side-by-side \href{https://musikisomorphie.github.io/Tera-MIND.html}{video visualizations} for more details.

\subsection*{Tera-MIND facilitated the translational application on unseen human brain samples.}
Following the translational aims of mouse brain studies in neuroscience, we conduct generalization experiments in which Tera-MIND was trained exclusively on mouse brain data (\textbf{Main} setting) and tested on the human brain healthy (Fig.~\ref{hbr} (a)) and glioblastoma (Fig.~\ref{hbr} (b)) samples~\footnote{\url{https://www.10xgenomics.com/datasets/xenium-human-brain-preview-data-1-standard}}. As shown in the gene panel (c) of Fig.~\ref{hbr}, we firstly determine the targeted genes that are present in both mouse brain atlases and human brain samples, which are then fed into the Tera-MIND model as the conditional input. After training with the patch-wise paired data from mouse brain atlases, we run the WSI generation using the gene expression arrays profiled on the healthy and glioblastoma human samples. Both qualitative (Fig.~\ref{hbr} (a, b)) and quantitative evaluations (Fig.~\ref{hbr} (d)) confirm that the performance of morphological simulation is generalizable from mouse to human samples, in terms of well-preserved tissue architecture and meaningful cellular correspondence to GT. Notably, Tera-MIND reported better PSNR and SSIM for the healthy brain sample than in the setting of Glioblastoma, suggesting a mild cross-species translational gap under non-physiological conditions.

When examining the gene expression patterns of GLUT pathway between healthy and glioblastoma cases, Fig.~\ref{hbr} (a) and (b) reveal distinct spatial distributions: \textit{Slc17a7} exhibits higher and more heterogeneous expression levels across the glioblastoma sample compared to the healthy sample, resulting in pronounced interaction signals between \textit{Slc17a6} and \textit{Slc17a7}. Conversely, both gene expression and their attention levels for the healthy sample remain homogeneous. This phenomenon can be attributed to the observation that glioma cells often release glutamate, leading to excitotoxic neuronal death and creating space for tumor expansion~\cite{sontheimer2008role}.

\section*{DISCUSSION}

In this study, we proposed a novel patch-based diffusion model Tera-MIND for high-fidelity simulation of mammalian brains. For three tera-scale mouse brain atlases, Tera-MIND achieved reproducible generation results and consistently identified spatial molecular interactions within GLUT and DOPA pathways. Beyond the comprehensive evaluation on murine brains, we established the \textit{in-silico} translational applicability of Tera-MIND by generalizing to previously unseen human brain samples, including both healthy and glioblastoma tissues. Methodologically, Tera-MIND offers an efficient and scalable generative framework for modeling whole (animal) organisms. From the viewpoint of clinical application, Tera-MIND thereby provides a novel \textit{in-silico} approach for comparative pathology~\cite{treuting2017comparative}, which enables a systematic and side-by-side analysis of murine and human tissues to assess the validity and translational relevance of animal models. Furthermore, Tera-MIND inherently facilitates \textit{in-silico} interventions on targeted genes of interest that drive (disease) morphological transitions~\cite{wuediting}. This capability allows for the exploration of `what if' causal questions for enhancing clinical treatment and diagnostics~\cite{wu2024towards}. Given the scope of this study, we plan to address the topic of simulated intervention in future work.

\subsection*{Limitations of the study}

In addition, we acknowledge the limitations of Tera-MIND. The modeling paradigm is fundamentally driven by the molecular-to-morphology spatial associations, which, while powerful, represent a simplified hypothesis of the complex neurological processes. With the rapid emergence of diverse spatial omics technologies~\cite{zhang2025spatial}, future work will incorporate orthogonal modalities, such as spatial proteomics and epigenomics, to provide additional layers of validation and further distinguish true functional spatial relationships.

In conclusion, the development of virtual replicas of entire biological organisms using GenAI, referred to as generative digital twins (GDTs~\cite{wu2024towards}) and exemplified by Tera-MIND in this study, opens new avenues for high-throughput biomedical simulation with minimal ethical concerns. This methodology offers a cost-effective and scalable alternative for preclinical testing of therapeutic strategies. Lastly, the potentially wide applicability of GDTs on human samples aligns with the principles of animal welfare by supporting the replacement, reduction, and refinement (3R) of animal use in laboratory research, providing a promising algorithmic tool for ethically responsible biomedical innovation.

\newpage







\section*{RESOURCE AVAILABILITY}


\subsection*{Lead contact}


Requests for further information and resources should be directed to and will be fulfilled by the lead contact, Jiqing Wu (Jiqing.Wu@unibas.ch).




\subsection*{Data and code availability}


\begin{itemize}
    \item The processed mouse brain data has been deposited at brain image library under the database identifier ace-lot-now and are publicly available as of the date of publication. All other data reported in this paper will be shared by the lead contact upon request.
    \item All original code has been deposited at Zenodo under the DOI 10.5281/zenodo.14826874 and is publicly available as of the date of publication.
    \item Any additional information required to reanalyze the data reported in this paper is available from the lead contact upon request.    
\end{itemize}

\section*{ACKNOWLEDGMENTS}


This research project is mainly supported through the core professorship funding of Prof. Dr. med.
Viktor Koelzer, Medical Co-Director and Professor of Experimental Pathology, University Hospital of Basel.

\section*{AUTHOR CONTRIBUTIONS}


Conceptualization, J.W. and V.H.K.; methodology, J.W.; investigation, J.W., I.B., V.H.K.; writing-–original draft, J.W., V.H.K.; writing-–review \& editing, J.W., I.B., V.H.K., Y.L., E.K.; funding acquisition, V.H.K.; supervision, V.H.K..

\section*{DECLARATION OF INTERESTS}


V.H.K. reports being an invited speaker for Sharing Progress in Cancer Care (SPCC) and Indica
Labs; advisory board of Takeda; and sponsored research agreements with Roche and IAG, all unrelated to the current study. V.H.K. is a
participant of patent applications in computational pathology outside ofthe submitted work.

\section*{DECLARATION OF GENERATIVE AI AND AI-ASSISTED TECHNOLOGIES}


During the preparation of this work, the author(s) used ChatGPT in order to polish the content. After using this tool or service, the author(s) reviewed and edited the content as needed and take(s) full responsibility for the content of the publication.

\section*{SUPPLEMENTAL INFORMATION INDEX}




\begin{description}
  \item Figures S1-S5 and their legends in a PDF
\end{description}

\newpage

\section*{MAIN FIGURE TITLES AND LEGENDS}



\begin{figure}
\centering
\includegraphics[width= 0.7\linewidth]{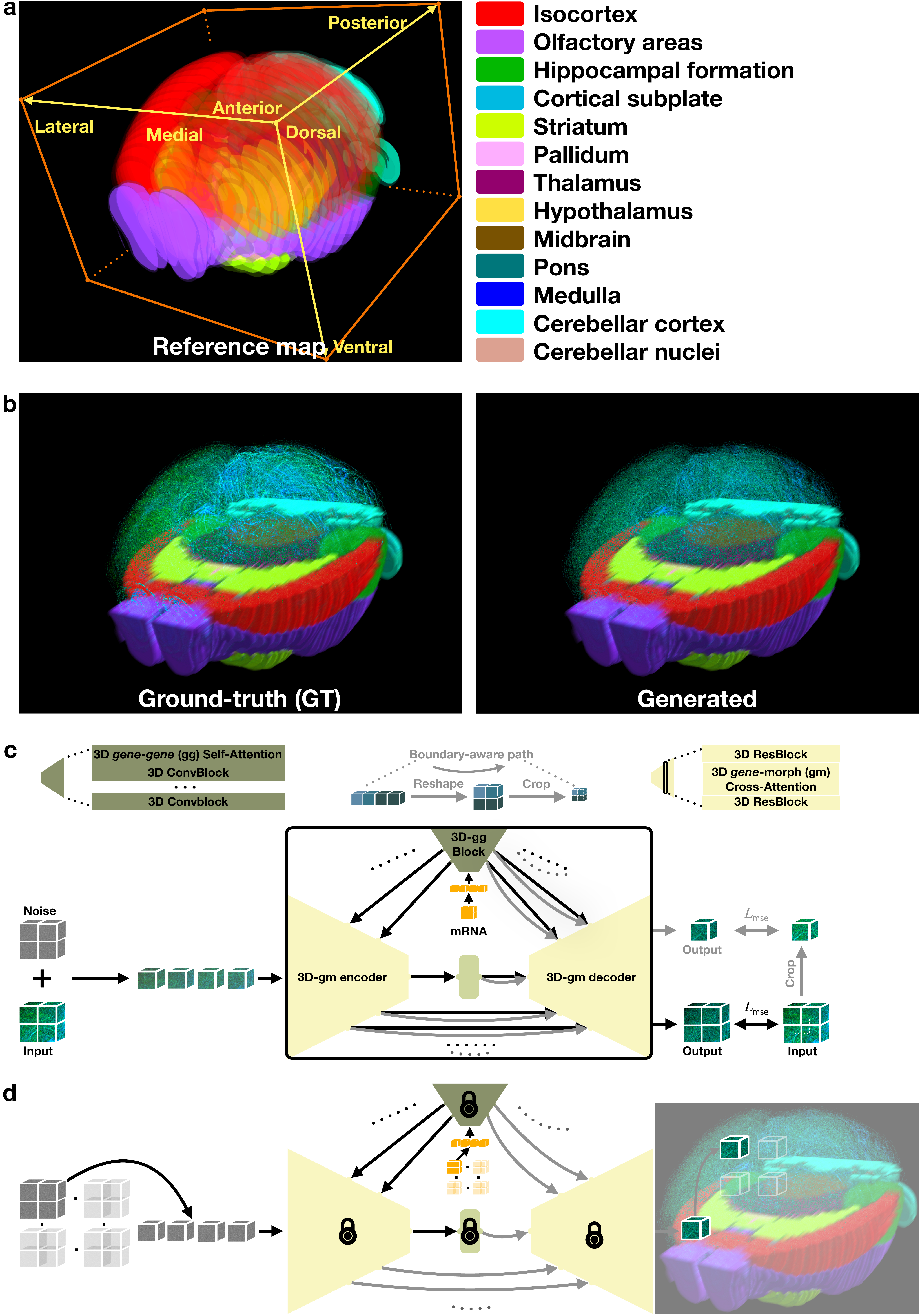}
\caption{\textbf{Overview of the Tera-MIND architecture and workflows for virtual organ simulation (Main result)}. \textbf{a}. The 3D reference map of major brain regions (left) alongside their corresponding labels (right) for guiding the navigation of brain structures.  \textbf{b}. The 3D visualization of the ground-truth (GT, left) and generated (right) mouse brain, as captured by a sequential stack of DAPI and PolyT-stained WSIs at the scale of $0.77\times 10^{12}$ voxels. The bottom part of generated and GT results overlay brain region maps, aiding in the navigation of the complex spatial brain architecture.  \textbf{c}. The conceptual illustration of patch-based diffusion model training. Here, we train Tera-MIND with noisy DAPI and PolyT bioimage patches so that the model learns to output clean patches guided by paired spatial mRNA array.  Apart from the standard reconstruction path (black arrows), we supply the boundary-aware path (gray arrows) to impose boundary consistency between generated neighboring patches.  \textbf{d}. The conceptual illustration of patch-wise generating virtual mouse brain using Tera-MIND. At this stage, we only run through the boundary-aware path to seamlessly generate the tera-scale mouse brain. }
\label{Fig1}
\end{figure}

\begin{figure}
\centering
\includegraphics[width= 0.7\linewidth]{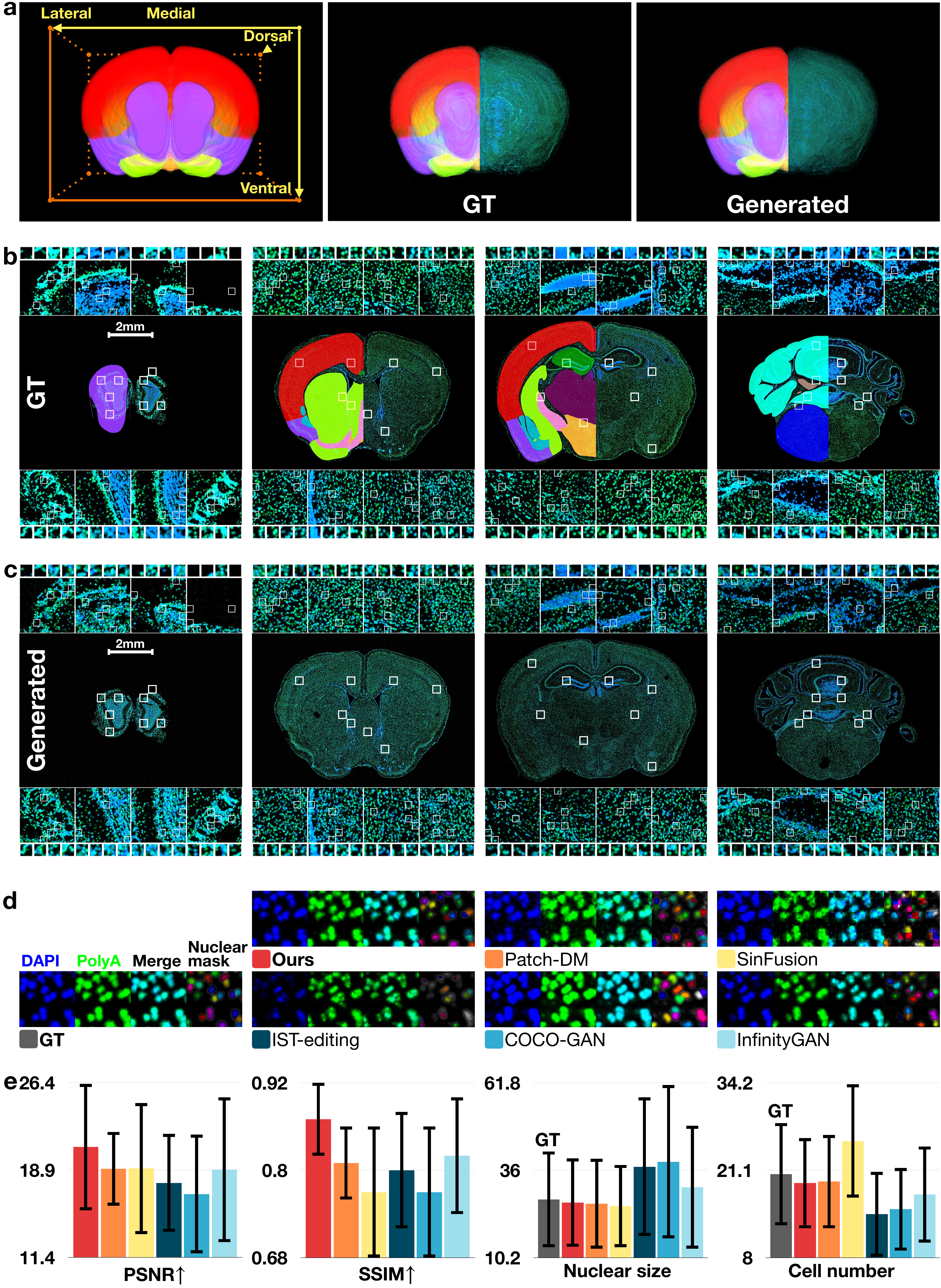}
\caption{\textbf{The visual and quantitative comparison of mouse brain generation (Main result)}. \textbf{a}. The 3D map of major brain regions (left), GT (middle), and the generated mouse brain (right)  of the atlas used in the primary analysis~\cite{yao2023high}. \textbf{b}. The cell- and region-level visualization of GT mouse brain WSIs in selected brain slices. For (a) and (b), the 3D/2D reference maps of brain regions are overlaid to aid in the navigation of brain structures. \textbf{c}. The cell- and region-level visualization of generated mouse brain WSIs. \textbf{d}. The visual comparison of exemplar cellular regions ($1024 \times 1024$) between Tera-MIND and state-of-the-art (SOTA) patch-based models. From left to right, the image stack displays individual \textcolor{blue}{DAPI}, \textcolor{green}{PolyT}, merged channels, and nuclear masks obtained using Cellpose~\cite{pachitariu2022cellpose}. \textbf{e}. Quantitative comparison of image generation quality between Tera-MIND and SOTA methods. This includes mean and standard deviation scores (error bar) for widely-used image quality metrics, such as PSNR and SSIM, alongside domain-specific scores like patch-based nuclear size and cell count. For a more in-depth analysis, we encourage readers to zoom in on the visual results for clearer comparisons.}
\label{Fig2}
\end{figure}

\begin{figure}
\centering
\includegraphics[width= 0.7\linewidth]{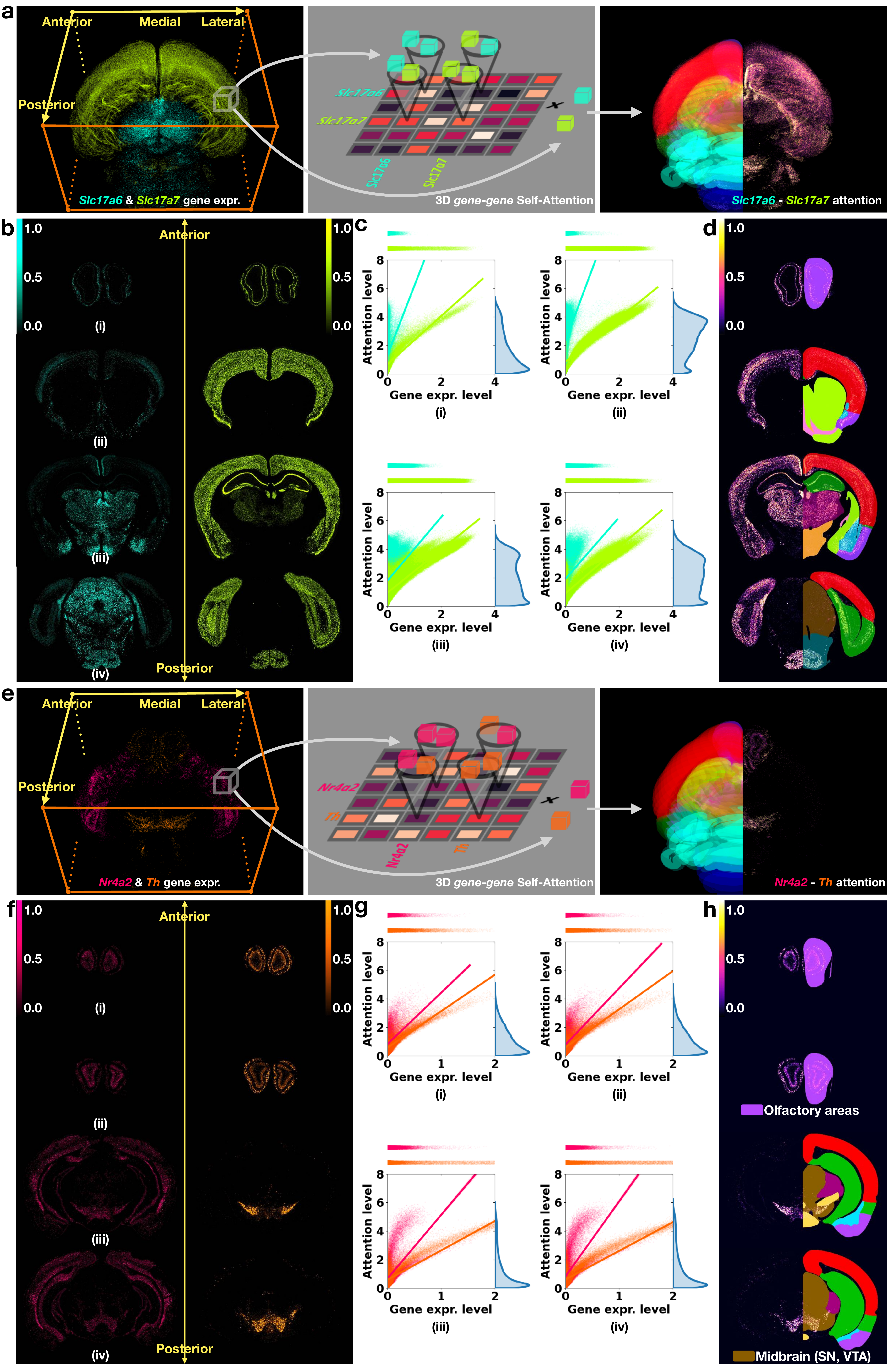}
\caption{\textbf{The visualization of \textit{gene}-\textit{gene} interactions for GLUT and DOPA neuronal systems (Main result)}. \textbf{a}. The 3D visualization of \textit{\textcolor{Slc17a6}{Slc17a6}} and \textit{\textcolor{Slc17a7}{Slc17a7}} gene expression (left), the conceptual illustration of 3D \textit{gene}-\textit{gene} attention layer (middle), and the 3D visualization of \textit{\textcolor{Slc17a6}{Slc17a6}}-\textit{\textcolor{Slc17a7}{Slc17a7}} attention (right). \textbf{b}. The WSI visualization of \textit{Slc17a6} and \textit{Slc17a7} gene expression (expr.). \textbf{c}. The linear regression analysis of gene expr. and attention level for both \textit{Slc17a6} and \textit{Slc17a7}. 
\textbf{d}. The WSI visualization of \textit{Slc17a6}-\textit{Slc17a7} attention level. \textbf{e}. The 3D visualization of \textit{\textcolor{Nr4a2}{Nr4a2}} and \textit{\textcolor{Th}{Th}} gene expression (left), the conceptual illustration of 3D \textit{gene}-\textit{gene} attention layer (middle), and the 3D visualization of \textit{\textcolor{Nr4a2}{Nr4a2}}-\textit{\textcolor{Th}{Th}} attention (right). \textbf{f}. The WSI visualization of \textit{Nr4a2} and \textit{Th} gene expression (expr.). \textbf{g}. The trending line showing the relationship between gene expr. versus attention level for both \textit{Nr4a2} and \textit{Th}. 
\textbf{h}. The WSI visualization of \textit{Nr4a2}-\textit{Th} attention level. Here, both SN and VTA are located towards the bottom of midbrain label. 
For a clearer visualization, gene expr. and attention levels are normalized to the range [0, 1] for both GLUT and DOPA pathways.}
\label{Fig3}
\end{figure}

\begin{figure}
\centering
\centerline{\includegraphics[width=0.7\linewidth]{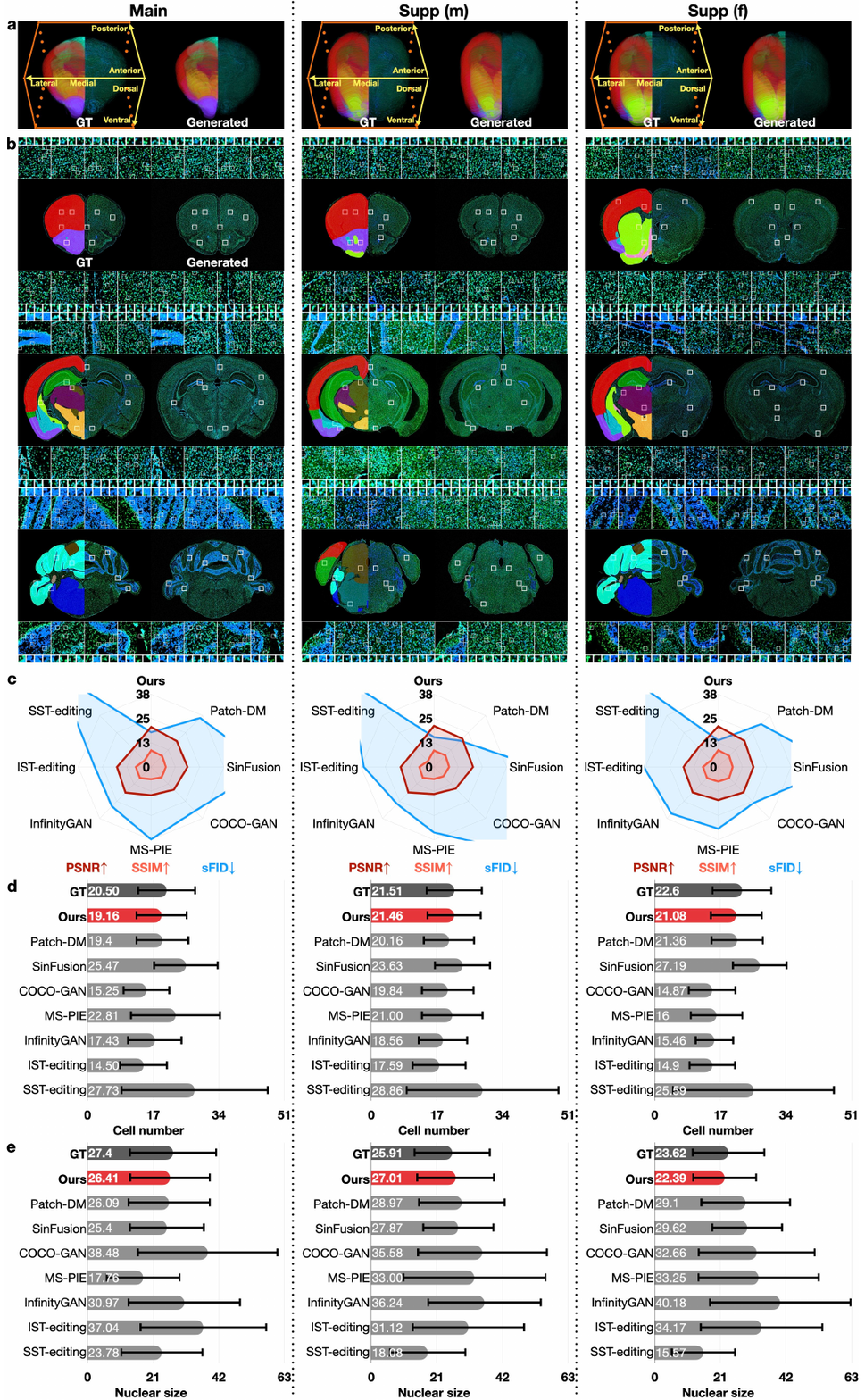}}
\caption{\textbf{The visual and quantitative comparison for the main and supp mouse brain results}. \textbf{a}. The holistic 3D  comparison between the generated and GT mouse brain, with the brain region map provided on the left for structural guidance. Note that all three generated mouse brains have the same $0.77 \times 10^{12}$ voxel scale. \textbf{b}. The cell- and region-level visualizations of WSIs for the side-by-side comparison between the generated and GT mouse brain. \textbf{c}. The quantitative comparison of generation quality between Tera-MIND and SOTA methods, as measured by PSNR, SSIM, and sFID. \textbf{d}. Compared to the GT results, the quantitative evaluation of cell numbers derived from Tera-MIND and SOTA methods. \textbf{e}. Compared to the GT results, the quantitative evaluation of nuclear size derived from Tera-MIND and SOTA methods.}
\label{Fig4}
\end{figure}

\begin{figure}[htp!]
\centering
\includegraphics[width= 0.7\linewidth]{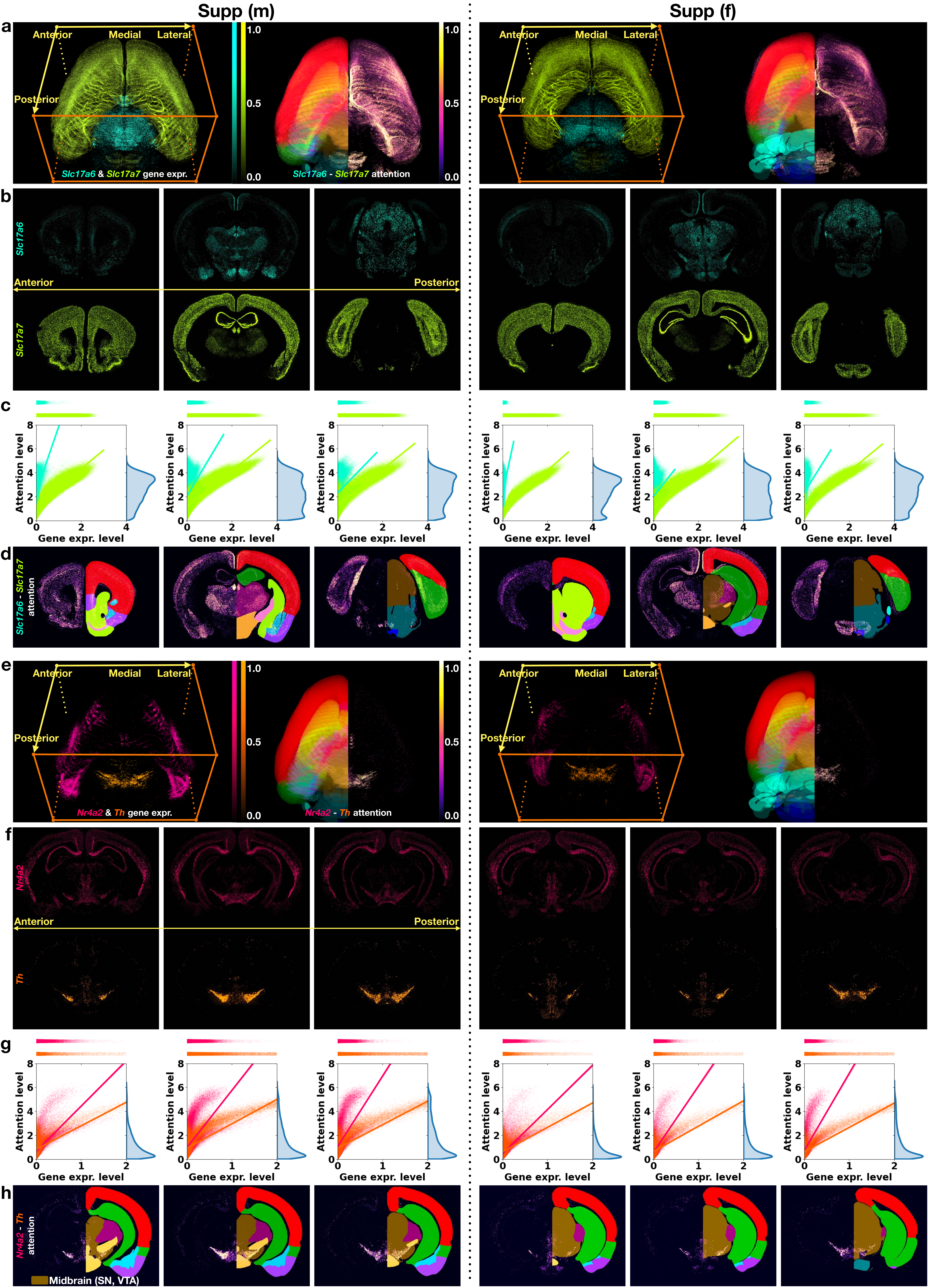}
\caption{\textbf{The visualization of \textit{gene}-\textit{gene} interaction for GLUT and DOPA neuronal systems (Supp (m) and (f) results)}. \textbf{a}. The 3D visualization of \textit{\textcolor{Slc17a6}{Slc17a6}} and \textit{\textcolor{Slc17a7}{Slc17a7}} gene expression (left) and the 3D visualization of \textit{\textcolor{Slc17a6}{Slc17a6}}-\textit{\textcolor{Slc17a7}{Slc17a7}} attention (right). \textbf{b}. The WSI visualization of \textit{Slc17a6} and \textit{Slc17a7} gene expression (expr.). \textbf{c}. The linear regression analysis of gene expr. and attention level for both \textit{Slc17a6} and \textit{Slc17a7}. \textbf{d}. The WSI visualization of \textit{Slc17a6}-\textit{Slc17a7} attention level.
\textbf{e}. The 3D visualization of \textit{\textcolor{Nr4a2}{Nr4a2}} and \textit{\textcolor{Th}{Th}} gene expression (left) and the 3D visualization of \textit{\textcolor{Nr4a2}{Nr4a2}}-\textit{\textcolor{Th}{Th}} attention (right). \textbf{f}. The WSI visualization of \textit{Nr4a2} and \textit{Th} gene expression (expr.). \textbf{g}. The linear regression analysis of gene expr. and attention level for both \textit{Nr4a2} and \textit{Th}. 
\textbf{h}. The WSI visualization of \textit{Nr4a2}-\textit{Th} attention level. Here, both SN and VTA are located towards the bottom of midbrain label. For a clearer visualization, gene expr. and attention levels are normalized to the range [0, 1] for both GLUT and DOPA pathways.}
\label{Fig5}
\end{figure}

\begin{figure}
\centering
\centerline{\includegraphics[width=0.7\linewidth]{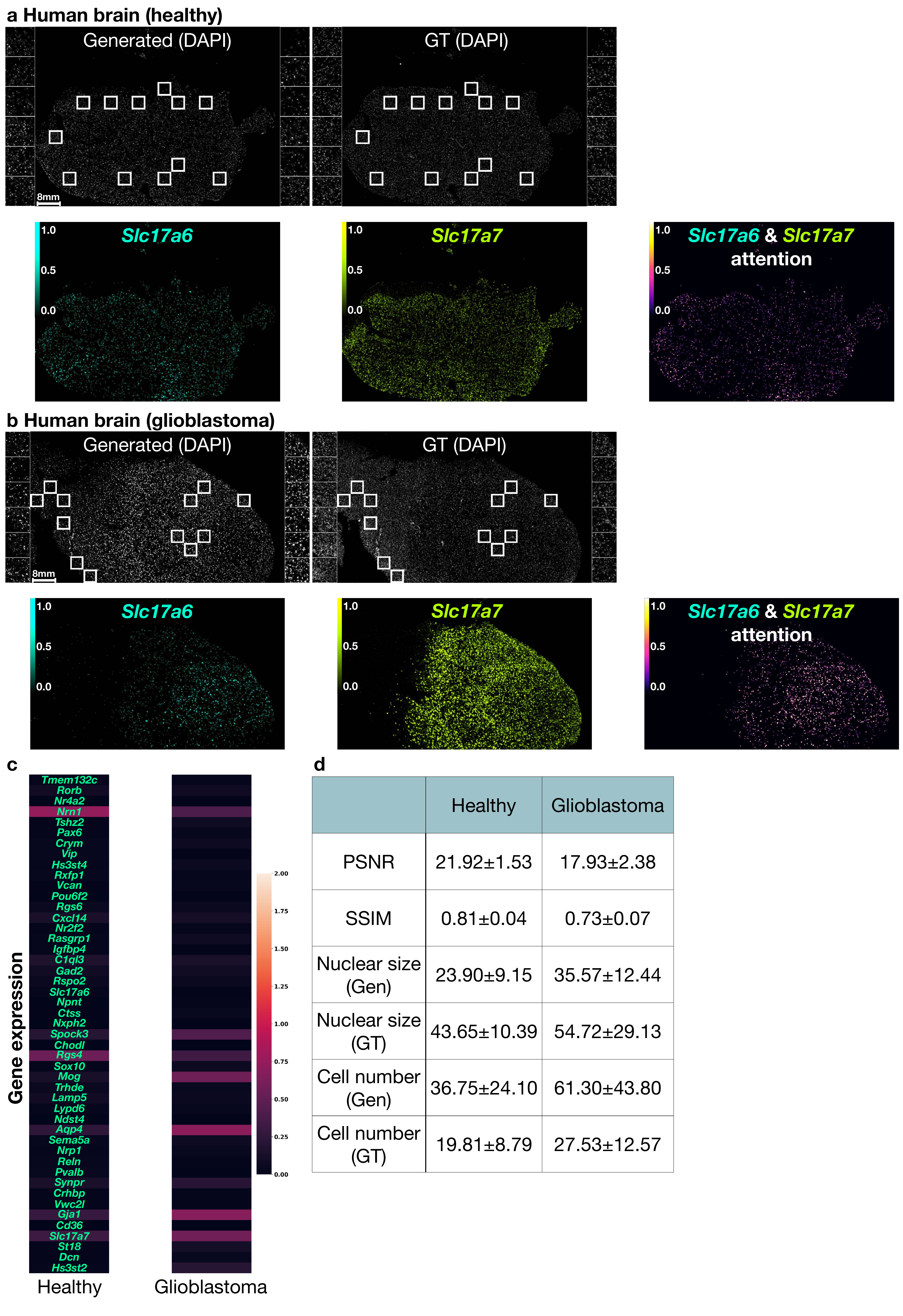}}
\caption{\footnotesize\textbf{The comprehensive generation, quantification, and gene expression statistics for the human brain experiments}. \textbf{a}. The side-by-side comparison between the generated (Gen) and ground-truth (GT) WSIs of the human brain (Healthy), with a $27136 \times 36864$ spatial resolution. The visualization of \textit{\textcolor{Slc17a6}{Slc17a6}} and \textit{\textcolor{Slc17a7}{Slc17a7}} gene expression and their spatial attention. \textbf{b}. The side-by-side comparison between the Gen and GT WSIs of the human brain (Glioblastoma), with a $23808 \times 39680$ spatial resolution. The visualization of \textit{\textcolor{Slc17a6}{Slc17a6}} and \textit{\textcolor{Slc17a7}{Slc17a7}} gene expression and their spatial attention. \textbf{c}. The side-by-side comparison of gene expression level (mean) for the human (Healthy) and human (Glioblastoma). \textbf{d}. The quantitative comparison of generation quality for the human (Healthy) and human (Glioblastoma), as measured by PSNR, SSIM, nuclear size, and cell number.
For a clearer visualization, gene expr. and attention levels are normalized to the range [0, 1] for the GLUT pathway.}
\label{hbr}
\end{figure}




\newpage

{
}

\bigskip


\newpage



\section*{STAR METHODS}
\label{sec:Methods}




\subsection*{Experimental model and study participant details}
Spatial transcriptomic data were obtained from three C57BL/6 mice (postnatal day 56; NCBI: txid10090): two adult females and one adult male. These whole-brain datasets were processed by the Allen Brain Institute and profiled using the MERSCOPE v1 platform.
Same as the primary analysis in~\cite{yao2023high} conducted for the female mouse (Sampleid: 638850), our main results used the same mouse brain atlas, whereas we reported supporting results on the remaining mouse brain atlases (Sampleid: 609882 and 609889).

For human samples, one healthy cortical section and one section containing glioblastoma multiforme were obtained from Avaden Biosciences. Spatial transcriptomic profiling of these sections was performed using the 10x Genomics Xenium platform.



\subsection*{Method details}


\subsubsection*{Spatial mRNA data as 3D `images'}
Biological processes are inherently spatial~\cite{bressan2023dawn}. The spatial molecular organization of the mammalian brain plays a decisive role in defining brain functionality and its dysfunction in disease. Recognizing the importance of spatial context, we leverage spatial mRNA readouts~\cite{yao2023high,zhang2023molecularly} as 3D `images', which are paired with corresponding 3D morphological bioimages. Unlike conventional data representations, such as 1D gene expression vector~\cite{cui2024scgpt,theodoris2023transfer}, text description~\cite{chen2024genept}, graph-based structure~\cite{jumper2021highly,abramson2024accurate}, or genomic sequence~\cite{nguyen2024sequence}, our simple and vision-driven data format naturally maintains the 3D localization of molecular features, facilitating the holistic modeling of intrinsic spatial organizations that underpins brain function and dysfunction.

\subsubsection*{Spatial paired data processing}
In prior studies~\cite{yao2023high,zhang2023molecularly}, three comprehensive and high-resolution brain atlases have been made publicly available~\footnote{\url{https://download.brainimagelibrary.org/aa/79/aa79b8ba5b3add56/}}. For each brain slice, there exists a large multiplex gene expression table and the associated morphological bioimages, as captured by DAPI-and PolyT-stained high-resolution WSIs.
Given the coordinate misalignment across available brain slices, we start the data processing by image registration. To maintain the original scale of raw data and for the sake of molecular data conversion, the registration-derived transformation matrix is required for our customized data processing. To this end, we employ the ABBA software~\cite{chiaruttini2024abba} to perform the brain atlas registration, which allows us to access individual $4 \times 4$ transformation matrices $\mathrm{S}$. By applying the inverse $\mathrm{S}^{-1}$ on the raw molecular and morphological data resp., we obtain the updated (large) gene expression table with transformed coordinates and the transformed WSI. For all brain slices, its transformed gene expression table is then converted to a sparse array, which has the same $73,216 \times 105,984$ spatial resolution as the corresponding transformed WSI. Specifically, we employ the same processing pipeline on all the available atlases and obtain three sets of gene expression 3D images and associated WSIs. After excluding the brain slices with corrupted image quality and post-processing, we keep 50 slices of paired data for each brain atlas. Note that the processed results have been carefully examined by domain experts to ensure the registration quality. We refer interested readers to Fig. S1-S3 (a) for more details. After consecutively stacking these gene expression arrays and WSIs resp., we obtain the whole tera-scale paired 3D `images', from which we can collect a large amount of small data pairs (\textit{e.g.}, $128 \times 128$) for running the model.

\subsubsection*{The proposed Tera-MIND}
To faithfully capture the molecular-to-morphology spatial associations, we introduce a 3D \textit{gene}-\textit{gene} block that learns the spatial gene expression embeddings, which are subsequently integrated into the 3D \textit{gene}-morph (gm) UNet to control the morphological reconstruction process (Fig.~\ref{Fig1} (c, d)). Inspired by DIT~\cite{peebles2023scalable,zheng2024open}, we apply newly designed spatial attention blocks to our model architecture.

\noindent\textbf{3D \textit{gene}-\textit{gene} (gg) self-attention}.  
Instead of 1D conditional embeddings learned from textual descriptions, we use the 3D-gg self-attention layer to process the 3D gene expression array used as the conditional prompt. Let $\mathbf{g} \in \mathbb{R}^{n \times d}$ be a $n$-plex 3D gene expression array with the spatial dimension $d = \mathrm{width} \times \mathrm{height} \times z$, where $z$ denotes the number of stacked slices from which the gene expression is extracted. Then, we have
\begin{equation}
\label{gg}
 \begin{aligned}
\mathbf{Q_g} &= \mathsf{RMSNorm}(\mathbf{g} \mathbf{W_q}), \; \mathbf{K_g} = \mathsf{RMSNorm}(\mathbf{g} \mathbf{W_k}), \\
\mathbf{Attn_g} &= \underbrace{\mathsf{Softmax}(\frac{\mathbf{Q_g}\mathbf{K_g^{T}}}{d}) \mathbf{g}}_{\mathbf{Attn_{gg}}} \mathbf{W_v}, \; \mathrm{where} \; \mathbf{V_g} =  \mathbf{g} \mathbf{W_v}.
\end{aligned}
\end{equation}
Here, $\mathbf{W_{\{q,k,v\}}} \in \mathbb{R}^{d \times d}$ are learnable weights and $\mathbf{Attn_{gg}}$ is the derived spatial \textit{gene}-\textit{gene} attention reported in Fig.~\ref{Fig3} and~\ref{Fig5}.
After stepwise 3D convolutional upscaling on the $\mathbf{Attn_g}$, we have the sequential of 3D gene features $\mathbf{g}_i \in \mathbb{R}^{d_i \times n_i}$ for $i = 1,\ldots, l$, with the increasing spatial resolutions by the magnitude of 2.

\noindent\textbf{3D \textit{gene}-morph (gm) cross-attention}. To integrate the learned $\mathbf{g}_i$ into the reconstruction of morphological images by 3D-gm UNet, we inject selected $\mathbf{g}_i$ to the corresponding layer of encoder and decoder, which process morphological presentations $\mathbf{m}_i \in \mathbb{R}^{d_i \times c_i}$ at the same spatial resolution $d_i$. First, we obtain the spatial adaptive coefficients $\mathbf{Scale_{g_{i}}}, \mathbf{Shift_{g_{i}}}, \mathbf{Gate_{g_{i}}} \in \mathbb{R}^{d_i}$ and gene embedding $\mathbf{Embed_{g_{i}}} \in \mathbb{R}^{d_i}$ by inputting the $\mathbf{g}_i$ to the $\mathsf{AdaLN}$~\cite{peebles2023scalable,zheng2024open} block containing $\mathsf{SiLU}$ and $\mathsf{Linear}$ layers. Then, we have  
\begin{equation}
\label{gm}
 \begin{aligned}
\mathbf{Ada_{m_i}} &= \mathsf{RMSNorm}(\mathbf{m_i}) \cdot (\mathbf{Scale_{g_{i}}} + 1) + \mathbf{Shift_{g_{i}}}, \\ 
\mathbf{P_{m_i}} &= \mathbf{m_i} +  \mathbf{Gate_{g_{i}}} \mathsf{Softmax}(\frac{\mathbf{Q_{Ada_{m_i}}}\mathbf{K_{Ada_{m_i}}^{T}}}{c_i}) \mathbf{V_{Embed_{g_i}}},
\end{aligned}
\end{equation}
where $\mathbf{Q_{Ada_{m_i}}}$ and $\mathbf{K_{Ada_{m_i}}}$ are query and key representations of $\mathbf{Ada_{m_i}}$, 
$\mathbf{V_{Embed_{g_i}}}$ is the value vector of $\mathbf{Embed_{g_{i}}}$.
For both attention blocks, we use the Pytorch implementation of flash attention-2~\cite{daoflashattention} to improve computational efficiency and reduce memory cost.

\noindent\textbf{Boundary-aware path}. Following the denoising diffusion framework~\cite{ho2020denoising}, we implement the forward process by gradually adding Gaussian noise to the 3D image patches $\mathbf{m}$ extracted from WSIs at timestep $t$, which gives us noisy image patches $\mathbf{m_t}$. Using 3D-gm UNet, we then parameterize the denoising function $\epsilon_\theta (\mathbf{m_t}, t)$ to reverse the process. 
Similar to~\cite{ding2023patched} and along with the standard denoising process, we supply an additional boundary-aware denoising path to impose the boundary consistency on the center-cropped patches. Therefore, our training objective is determined to be $\|\epsilon_\theta (\mathbf{m_t^1}, t) - \epsilon_1 \|^2 + \|\epsilon_\theta (\mathbf{m_t^2}, t) - \epsilon_2 \|^2$, where $\mathbf{m_t^1}$ is the stack of small patches extracted from the input data and $\mathbf{m_t^2}$ is the center-cropped patch (See also Fig.~\ref{Fig1} (c)). During the inference, we only use the boundary-aware path to generate image patches without stitching artifacts.

\noindent\textbf{Model training}.
Among available mouse brain atlases, previous studies~\cite{yao2023high,zhang2023molecularly} reported the primary results on the P56 female mouse brain (\textbf{Main}) including all the major brain regions (Fig.~\ref{Fig1} (b)), while the additional P56 male and female mouse brains (\textbf{Supp (m), (f)}) that preserve most of the brain regions (Fig.~\ref{Fig3} (a)) are used for supporting analyses. Following this experimental setup and to avoid overfitting, we report our primary generation result on the main mouse brain while taking the supp (m) and (f) mouse brains as the training data. Given the heterogeneous gene lists profiled on three atlases, we filter out the genes that are present across three instances and obtain the 279-plex gene expression array for training (Fig. S1 (e)). In addition to the main results reported in Fig.~\ref{Fig1},~\ref{Fig2}, we conduct (unseen) generation experiments on the supporting mouse brains for the sake of reproducibility and consistency. Since supp (m) and (f) are processed using the same protocol with the identical $500$-plex gene list (Fig. S2-3 (e)), we take one `supp' brain for training and the other unseen `supp' brain for testing, and vice versa. Given the small distribution shift between supporting brain instances, this setup thus complements the challenging main experiment, which presents a large distribution gap between the training (supp (m, f)) and testing (main) data. 

\noindent\textbf{Model inference}.
After completing the model training, we run the tera-scale brain generation in a patch-wise manner. To ensure both computational feasibility and generation fidelity, we employ the Denoising Diffusion Implicit Models (DDIM)~\cite{songdenoising} approach for accelerated sampling and high-quality image synthesis. Given the significant GPU memory constraints associated with storing an entire mouse brain, the collection of generated patches is instead temporarily offloaded to the hard drive during each sampling step. These intermediate results are then utilized for the subsequent step. The patch-wise generation framework is inherently parallelizable, enabling efficient resource utilization. On a single NVIDIA A100 DGX system, the entire generation process for a tera-scale mouse brain can be completed within seven days. This demonstrates the scalability and practicality of Tera-MIND for handling large-scale 3D biomedical datasets.

\subsubsection*{Ablative studies}
To assess the impact of key design choices in Tera-MIND, we discuss ablative results based on three critical factors: the number of slices used to extract 3D training data pairs, the spatial resolution of image patches, and the number of sampling steps. In line with the previous evaluations, both general metrics (PSNR and SSIM) and domain-specific metrics (Nuclear size and Cell number) are reported for a more comprehensive analysis.

For all three brain instances, optimal quantitative performance has been achieved when the slice number was set to 2. This is supported by the fact that neighboring slices are separated by a relatively large interval, ranging from $100\mu m$ to $200\mu m$~\cite{yao2023high}. When compared to the neighboring pixels, which have a physical size of $0.108\mu m$ per pixel, the morphological context extracted from more neighboring slices is very likely saturated, as illustrated in Fig. S5 (a). Besides, in comparison to the 2D image training paradigm when the slice number equals 1, our proposed 3D modeling can further learn the cross-slice interaction patterns between neighboring slices. For instance, Fig. S4 highlights the consistent spatial \textit{Nr4a2}-\textit{Th} interaction patterns pinpointed in the very SN and VTA regions across three simulated mouse brains. These results substantiate the superiority of the proposed 3D modeling for the whole biological entity.    

Additionally, we explore the effect of varying image patch resolution, ranging from $64 \times 64$ to $256 \times 256$ (Fig. S5 (b)). Here, we consider the entire patch size of the input data, rather than using an intermediate cropped patch size.  Eventually, we determine that the resolution of $128 \times 128$ yields the best performance. This choice aligns with the previous Patch-DM study\cite{ding2023patched}, where the whole patch resolution was identified as 128 by default.

Finally, we examined the impact of the DDIM sampling step, weighing the trade-off between computational cost and performance. As shown in Fig. S5 (c), the optimal sampling step for Tera-MIND is determined to be 15. Beyond this point, additional steps lead to saturated improvements in performance, underscoring the cost-effectiveness of this choice.

\end{document}